\documentclass[10pt,twocolumn,letterpaper]{article}

\usepackage{cvpr}
\usepackage{times}
\usepackage{epsfig}
\usepackage{graphicx}
\usepackage{amsmath}
\usepackage{amssymb}

\usepackage[breaklinks=true,bookmarks=false]{hyperref}

\cvprfinalcopy % *** Uncomment this line for the final submission

\def\cvprPaperID{****} % *** Enter the CVPR Paper ID here

\begin{document}

%%%%%%%%% TITLE
\title{Online Illumination Invariant Moving Object Detection by\\Generative Neural Network}

\author{Fateme Bahri \qquad Moein Shakeri \qquad Nilanjan Ray\\
Department of Computing Science, University of Alberta\\
Edmonton, Alberta, Canada\\ 
{\tt\small \{fbahri, shakeri, nray1\} @ualberta.ca}
}

\maketitle
%\thispagestyle{empty}

%%%%%%%%% ABSTRACT
\begin{abstract}
Moving object detection (MOD) is a significant problem in computer vision that has many real world applications. Different categories of methods have been proposed to solve MOD. One of the challenges is to separate moving objects from illumination changes and shadows that are present in most real world videos. State-of-the-art methods that can handle illumination changes and shadows work in a batch mode; thus, these methods are not suitable for long video sequences or real-time applications. In this paper, we propose an extension of a state-of-the-art batch MOD method (ILISD) \cite{shakeri2017moving} to an online/incremental MOD using unsupervised and generative neural networks, which use illumination invariant image representations. For each image in a sequence, we use a low-dimensional representation of a background image by a neural network and then based on the illumination invariant representation, decompose the foreground image into: illumination change and moving objects. Optimization is performed by stochastic gradient descent in an end-to-end and unsupervised fashion. Our algorithm can work in both batch and online modes. In the batch mode, like other batch methods, optimizer uses all the images. In online mode, images can be incrementally fed into the optimizer. Based on our experimental evaluation on benchmark image sequences, both the online and the batch modes of our algorithm achieve state-of-the-art accuracy on most data sets.
\end{abstract}

%%%%%%%%% BODY TEXT

\section{Introduction}

\begin{figure}
\begin{center}
\includegraphics[width=\columnwidth]{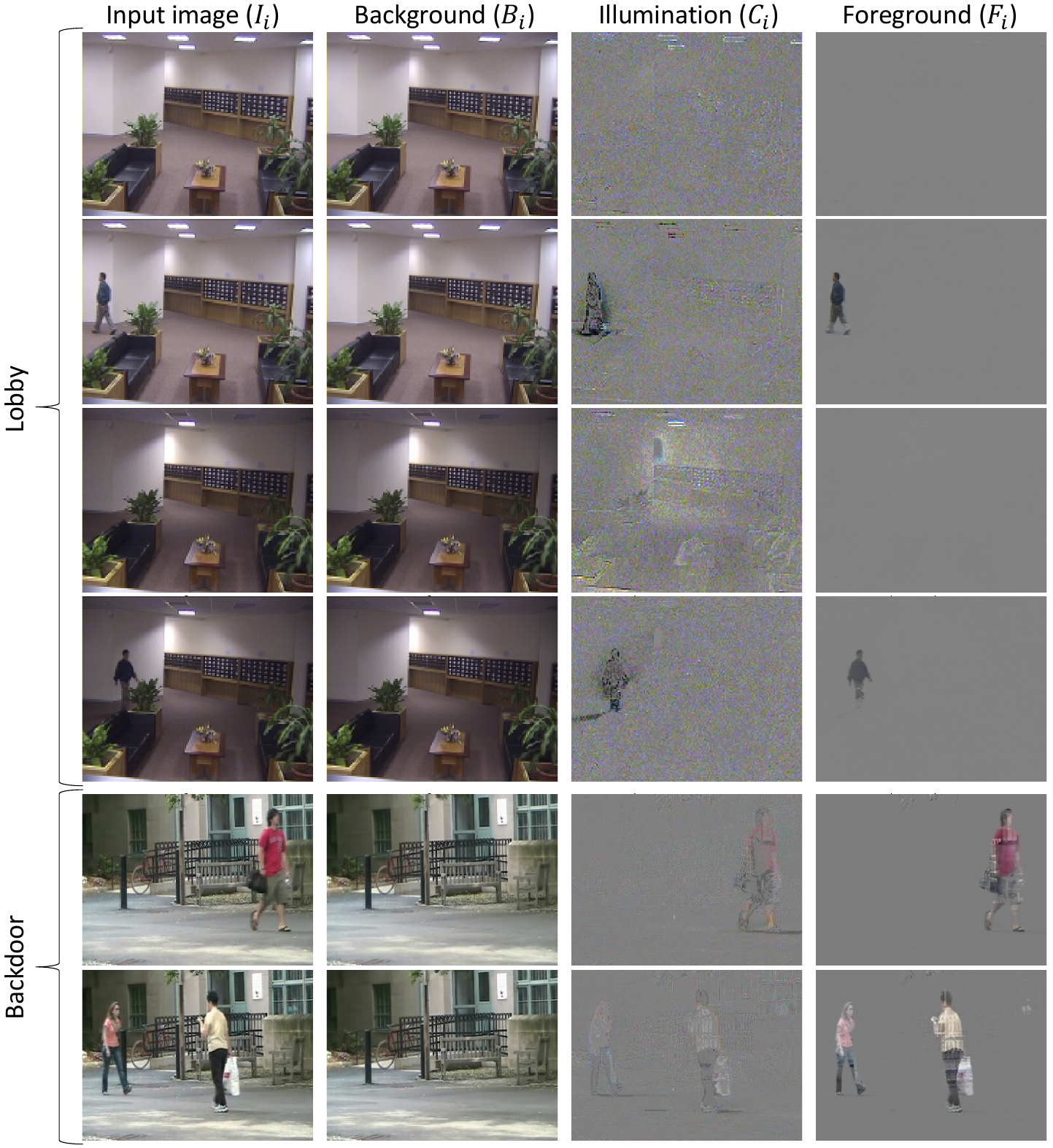}
\caption{Proposed method (NUMOD) decomposes input image, $I_i$, into background image, $B_i$, illumination changes and shadows, $C_i$ and foreground moving objects $F_i$. Examples are shown on two benchmark sequences ``Lobby'' and ``Backdoor.''}
\label{fig:decomposition}
\end{center}
\end{figure}

Moving object detection (MOD), which tries to separate moving objects out of a sequence of images, is one of the fundamental tasks in computer vision that has various applications, such as video surveillance in airports, residential areas, shopping malls \cite{tian2012robust,brutzer2011evaluation}, traffic monitoring \cite{sen2004robust,chen2014advanced}, object tracking \cite{saravanakumar2010multiple,suresh2014efficient} and gesture recognition in human machine interaction \cite{tsai2013algorithm}. Different types of methods have been proposed for MOD. However, many of these methods are vulnerable to illumination changes.

For resolving MOD, traditional statistical methods model pixel intensity of background and compare pixels of each frame to it to find the foreground. A single Gaussian (SG) \cite{wren1997pfinder} is the simplest method in this category that models pixel intensity. Later, Gaussian Mixture Models (GMM) \cite{stauffer1999adaptive} was proposed that models pixels by a number of Gaussians and uses an online approximation to update the model. GMM is an online algorithm and can also model dynamic background to some degree. Kernel Density Estimation (KDE) \cite{elgammal2000non} is another statistical method for MOD. Although these methods are computationally efficient, they cannot handle illumination changes and shadows.

State-of-the-art methods for MOD are based on low-rank and sparse (LRS) matrix decomposition. These methods use the fact that background pixels are linearly correlated to each other temporally in a sequence of images and usually apply a variant of Principal Component Analysis (PCA) on a sequence of images to obtain a low-rank representation for the background and sparse outliers representing moving objects. 

PCP method \cite{candes2011robust} solved the matrix decomposition problem by minimizing a combination of the nuclear norm of the low-rank matrix and $L_1$-norm of the sparse matrix. Zhou \textit{et al.} \cite{zhou2011godec} accelerated the decomposition in their proposed algorithm GoDec and its faster variant Semi-Soft GoDec (SSGoDec). Wang \textit{et al.} \cite{wang2012probabilistic} proposed Probabilistic Robust Matrix Factorization (PRMF) that uses a Laplace error and a Gaussian prior. A group of the LRS methods made use of connectivity constraint on moving objects \cite{wang2013bayesian, zhou2013moving, liu2015background}. In \cite{wang2013bayesian} a Bayesian robust matrix factorization (BRMF) model is proposed. Its extension, Markov BRMF (MBRMF), assumes outliers in foreground form groups with spatial or temporal proximity by placing a first-order Markov random field. 

Another method called Decolor \cite{zhou2013moving} assumes foreground objects form small contiguous regions and incorporates prior knowledge of contiguity in detecting outliers using Markov Random Fields. Liu \textit{et al.} \cite{liu2015background} proposed Low-rank and Structured sparse Decomposition (LSD) framework. LSD considers spatial information in sparse outliers and model the foreground by structured sparsity-inducing norms. It also uses a motion saliency map to remove background motion from foreground candidates, thus, it can tolerate some sudden background variations. In general, for scenes with moderate illumination changes, LRS methods can handle illumination changes. However, with the presence of significant illumination changes and shadows, LRS methods fail to separate moving objects from illumination changes \cite{shakeri2017moving}.

Recently, a new method called ILISD \cite{shakeri2017moving} has been proposed, which is based on low-rank and sparse decomposition. In addition, ILISD incorporates prior knowledge about illumination and is been able to distinguish between illumination changes and real foreground changes. The drawback of this method is that it can only work in a batch mode. For continuous monitoring tasks and very long sequences, an online (incremental) method is required. When number of images in the sequence increases, memory storage and computations grow significantly for ILISD.

Another group of MOD methods works in an online/incremental manner. Most of these method are based on robust PCA. GRASTA is an online incremental gradient descent algorithm which estimates robust subspace from subsampled data \cite{he2012incremental}. OPRMF is the online extension of PRMF which uses expectation-maximization algorithm which can be updated incrementally \cite{wang2012probabilistic}. OR-PCA proposes an online robust PCA method. To process the frames incrementally, it uses multiplication of the subspace basis and coefficients instead of nuclear norm and updates the basis for the new frame \cite{feng2013online}. COROLA \cite{shakeri2016corola} method uses a sequential low-rank approximation on a fixed window of images. Moreover, it considers outliers as small contiguous regions like the Decolor method. Although these methods have been applied successfully in real-time moving object detection, they fail to provide satisfactory results, when significant illumination changes are present.

Inspired by the significant success of deep neural networks in computer vision, several MOD methods based on deep networks have been proposed recently \cite{babaee2018deep, braham2016deep, sakkos2017end, wang2012probabilistic, zeng2018background, yang2018deep, chen2017pixel}. However, most of the deep learning-based methods need supervised training with pixel-wise ground-truth masks of moving objects. Since, annotating images for every pixel is an expensive task and not always practical for the real-world environment, unsupervised optimization is preferred. To the best of our knowledge, the only unsupervised neural network based method, BEN-BLN, is proposed in \cite{xu2014dynamic}. In this work, a stack of autoencoders,called Background Extraction Network (BEN), is used to estimate a background model and then a second autoencoder, called Background Learning Network (BLN), is in charge of enhancing the background. Also an online extension of this method is proposed, BEN-BLN-Online, that trains the network on a first batch of images and then finetunes it for streams of input images. Even though BEN-BLN model builds a non-linear background model, it is unable to accommodate illumination changes.
% Generative Network for Invariant Object Detection (GNIOD)
%Neural Online Moving Object Detection (NOMOD)
%Neural Online Invariant Moving Object Detection (NOIMOD)

In this paper, we propose an end-to-end framework called Neural Unsupervised Moving Object Detection (NUMOD) following the principles of the batch method, ILISD. Because of the parameterization via generative neural network, NUMOD can work \textbf{both in the online and in the batch mode.} NUMOD's goal is decomposing each frame into three parts: background, illumination changes and moving objects that we are interested in. It uses a fully connected generative neural network to generate a background model by finding a low-dimensional manifold for the background of the image sequence. For each image, after subtracting the background, the sparse outliers are remained. The sparse outliers include not only moving objects, but also moving shadows and illumination changes. To distinguish between them, NUMOD uses an illumination invariant representation of the images as a prior knowledge that is robust against illumination changes and shadows. This representation has been successfully used in ILISD \cite{shakeri2017moving}. NUMOD adds some constraints to the loss function based on this invariant representations that enables it to decompose the sparse part of each image into illumination and foreground.

Fig.~\ref{fig:decomposition} shows example decomposition of input images obtained by NUMOD. This qualitative results illustrate how NUMOD decomposes an input image, $I_i$, to background image, $B_i$, illumination changes and moving shadows, $C_i$, and detected foreground objects, $F_i$. The first four rows are selected from `Lobby'' sequence, which is an indoor scene with illumination changes. It can be seen that background images capture some part of the illumination changes, but most of the illumination changes and shadows are captured in $C_i$. The foreground image, $F_i$, is free of shadows and illumination changes. The last two rows are from ``Backdoor'' sequence, which is an outdoor scene with moving shadows. Again, we observe that moving shadows and illumination changes are separated from moving objects. These qualitative results show capability of our method in handling illumination changes and shadows. We also perform quantitative experiments.

NUMOD's advantages can be summarized as follows. First, it can work in both batch and online modes of operation. Second, it uses prior knowledge to overcome illumination challenge in an end-to-end neural network framework. Third, unlike other neural network based methods, it trains in an unsupervised way, without requiring expensive pixel-wise ground-truth masks. In a nutshell, the main contribution of NUMOD is that it is an online method, which has excellent capability of handling illumination changes.

The rest of the paper is structured as follow: Section 2 explains the illumination invariant representation we use in NUMOD. In section 3, NUMOD methodology and framework are described. We report our experimental results in section 4. Finally, Section 5 provides conclusions and an outline of future work.

\section{Illumination invariance prior knowledge}
Many methods have been proposed for shadow free images and illumination invariant images. One of the well known methods is proposed by Finlayson \textit{et al.} \cite{finlayson2006removal}, which shows that for an RGB image under illumination changes, the 2D log-chromaticity vector for a color surface moves along a straight line in the scatter-plot of $log(R/G)$ vs. $log(B/G).$ The direction of this line, $e$, is same for different surfaces. If we project all the chromaticities on a line orthogonal to $e$ then all the points of the same surface, independent of the illumination, will be projected to the same point. This model provides a shadow free image useful for distinguishing between real moving objects and illumination changes. 

However, \cite{finlayson2006removal} has some initial assumptions, which do not always hold in real data sets. The model assumes that the scene's illumination is Planckian, the camera sensors are narrow-band and the image surfaces have Lambertian reflectance. When these assumptions are not correct, chromaticities of an image's surface do not move along a straight line. 

To overcome this problem, we use Wiener filter to get illumination invariant features of images while preserving their structural information \cite{chen2011illumination}. In \cite{chen2011illumination} Wiener filter is used to separate illumination and reflectance of an image across the whole frequency spectrum. The advantage of this method is that unlike other methods, it considers low frequency part of spectrum as well. Consequently, it preserves features at every frequency. 

Let $I_i, i=1,2,...,n$ be an input image sequence. We combine Finlayson \textit{et al.}'s shadow free images and illumination invariant images extracted by Wiener filter by a simple averaging as proposed in \cite{shakeri2016illumination}. The following function denotes transformation of an input image into an illumination invariant representation:
\begin{equation} \label{eq:2}
I_i^{inv} = \Psi (I_i).
\end{equation}

\begin{figure}
\begin{center}
\includegraphics[width=\columnwidth]{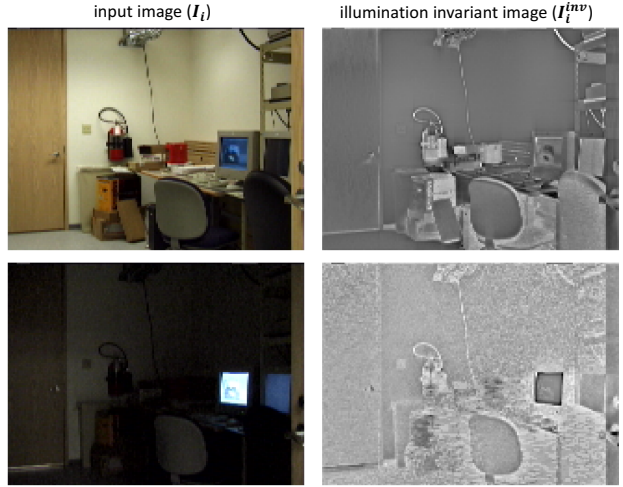}
\caption{Left column shows two input images, $I_i$, from ``LightSwitch'' sequence representing illumination changes; Right column shows corresponding illumination invariant images, $I_i^{inv}$.}
\label{invariant}
\end{center}
\end{figure}

Fig. \ref{invariant} shows two input images $I_i$ of a same scene under illumination changes and their respective invariant representations $I_i^{inv}.$

\section{Proposed method: NUMOD}
Let $I_i\in R^{m}, i=1,2,...,n$ be a sequence of vectorized RGB input images. Our method decomposes $I_i$ into three images: a background image $B_i$, an image $C_i$ representing illumination change and another image $F_i$ denoting moving objects as follows:
\begin{equation} \label{eq:1}
I_i = B_i + C_i + F_i.
\end{equation}

\begin{figure*}
\begin{center}
\includegraphics[width=\textwidth]{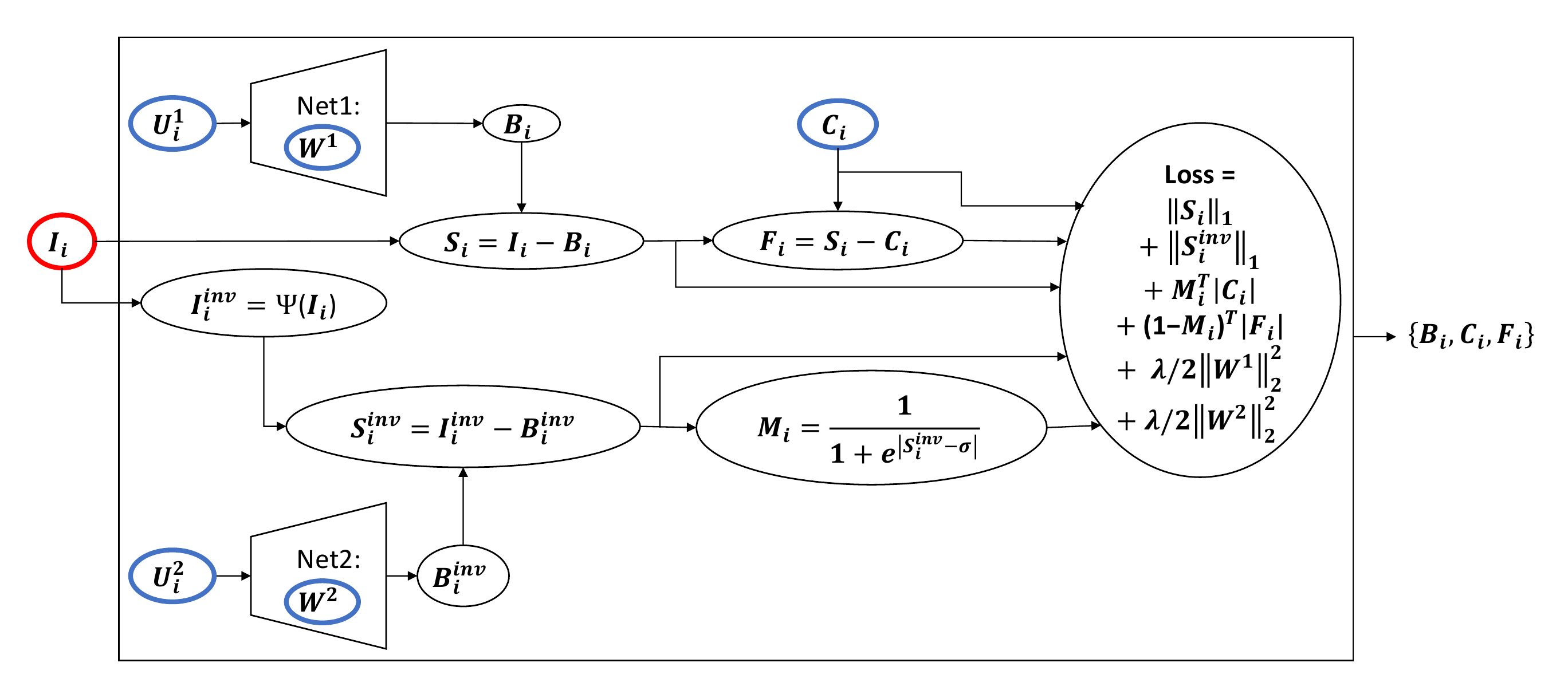}
\caption{The graph shows the flow of the computations in our framework for the $i^{\text{th}}$ image frame. Input image $I_i$ is shown in red outline. Net1 and Net2 are the two generative neural networks with parameters $W^1$ and $W^2$, respectively. Optimizable variables $\{U_i^1, U_i^2, W^1, W^2, C_i\}$ are shown in blue outline. Output is the triplet $\{B_i, C_i, F_i\}$ with the relation: $I_i=B_i+C_i+F_i$.}
\label{graph}
\end{center}
\end{figure*}

The flow of computations for an input frame is shown in Fig. \ref{graph}. In summary, the background image $B_i$ is generated using a fully connected generative network with a low dimensional input vector $U_i^1$. We also decompose the remaining sparse vector ($I_i-B_i$) into illumination changes $C_i$ and foreground moving objects $F_i$ by applying illumination invariant constraints. We will explain details of our algorithm in the following sections.

\subsection{Generative network architecture} 

\begin{figure}
\begin{center}
\includegraphics[width=\columnwidth]{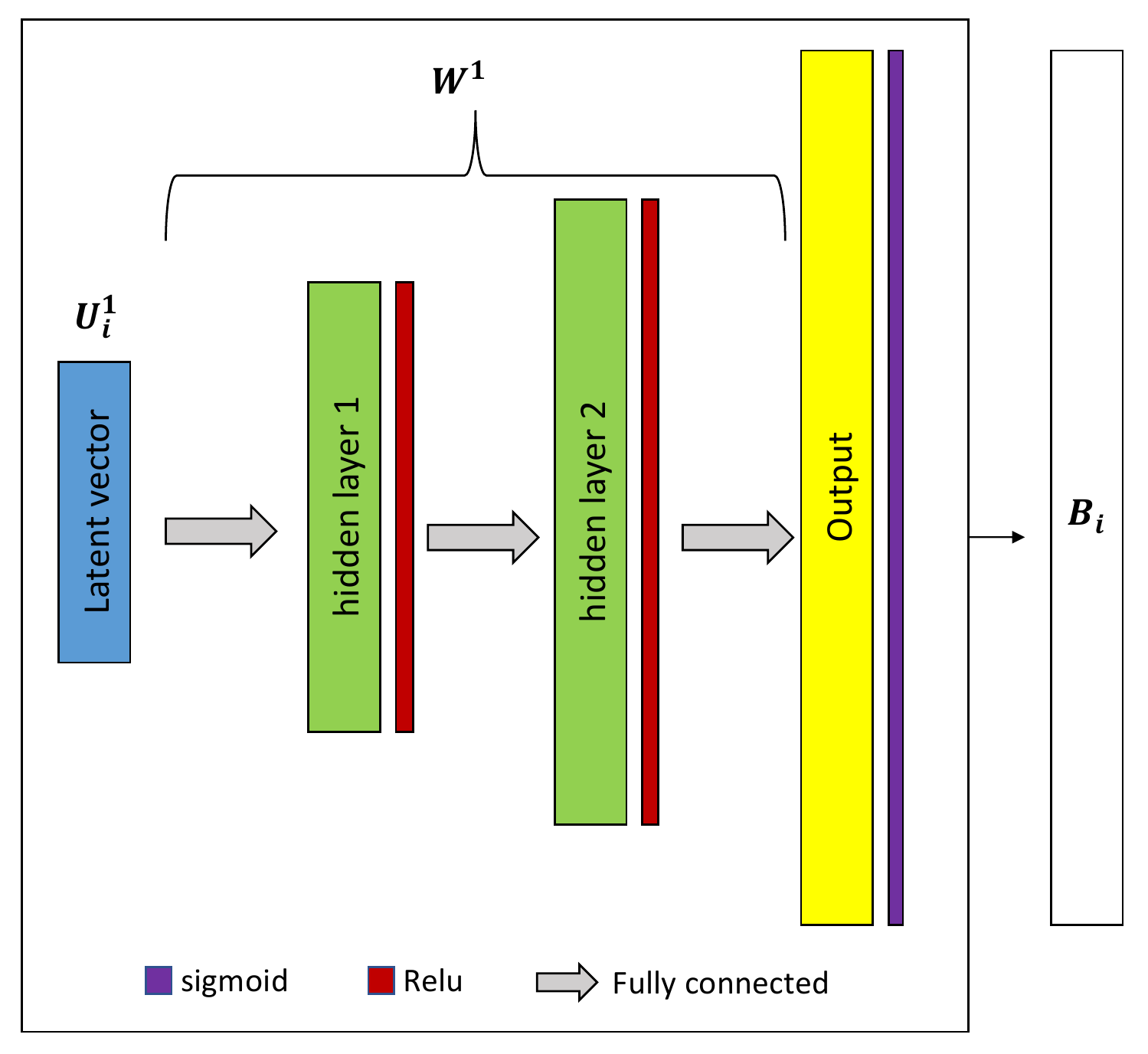}
\caption{Generative fully connected network: Net1}
\label{GFCN}
\end{center}
\end{figure}

There are two Generative Fully Connected Networks (GFCN) in NUMOD: Net1 and Net2 (Fig.~\ref{graph}). Net1 is in charge of estimating background image, $B_i$, from the input image $I_i$ and Net2 generates background image $B^{inv}_i$ for the illumination invariant image $I^{inv}_i$. These two networks have the exact same architecture which is shown in Fig.~\ref{GFCN}.

Input to GFCN is an optimizable low-dimensional latent vector ($U^1_i$ and $U^2_i$ in Fig.~\ref{graph}). After that there are two fully connected hidden layers each followed by ReLU non-linearity. The second hidden layer is fully connected to the output layer which is followed by the sigmoid function. The reason to use the sigmoid function at the last layer is to limit background values between zero and one.

A loss term (\ref{Reconst_loss}) imposes that the output of GFCN be similar to the current input frame. GFCN is similar to the decoder part of an autoencoder. In an autoencoder, the low-dimensional latent code is learned by the encoder, whereas in GFCN, it is a free parameter that can be optimized and is the input to the network. During training, this latent vector learns a low-dimensional manifold of the input distribution. If no further constraint is applied, the network will learn a naive identity function. Hence by restricting the capacity of the network, and by limiting number of hidden units in the hidden layers, the network is able to extract the most salient features of the data and gets a structure of the data distribution~\cite{goodfellow2016deep}. In our problem, since images of the sequence are temporally correlated to each other, GFCN is able to learn a background model of the images and output of the network is the background image.

The $L_{reconst}$ loss term, responsible for constructing background images of $I_i$ and $I^{inv}_i$, is as follows:
\begin{equation} \label{Reconst_loss}
L_{reconst} = \sum_i\lVert I_i - B_i\rVert_1 +\sum_i \lVert I^{inv}_i - B^{inv}_i\rVert_1,
\end{equation}
where $B_i$ and $B^{inv}_i$ are outputs of Net1 and Net2, respectively. We use $L_1$-norm instead of $L_2$-norm in $L_{reconst}$ to encourage sparsity of the sparse remainder of images \cite{candes2008enhancing}. 

\subsection{Illumination and foreground decomposition}
NUMOD is a unified framework shown in Fig. \ref{graph}. In the previous section, we explained how Net1 and Net2 generate backgrounds for an input image and an invariant image, respectively. Subtracting the background from an input image gives the sparse outliers of that frame (\ref{eq:4}). $S_i$ includes illumination changes and foreground moving objects:
\begin{equation} \label{eq:4}
\begin{split}
& S_i =  I_i - B_i \\
& S^{inv}_i = I^{inv}_i - B^{inv}_i.
\end{split}
\end{equation}
As explained earlier, using Net2, We decompose an illumination invariant image $I^{inv}_i$ into invariant background image $B^{inv}_i$ and invariant sparse foreground moving objects $S^{inv}_i$. Since, we assume that $I^{inv}_i$ is independent of illumination changes, we can use its sparse part $S^{inv}_i$ as a map for actual moving objects as follows \cite{shakeri2017moving}:
\begin{equation} \label{map}
 M_i = \frac {1} {1+e^{-(\mid S^{inv}_i-\sigma \mid)}}, 
\end{equation}
where $\sigma$ is the standard deviation of pixels in $I^{inv}$ and Sigmoid function normalizes $M_i$ values between zero and one.

Prior map $M_i$ is used to apply constraints to separate real foreground changes from illumination changes and moving shadows. One of the initial assumptions is that in ideal circumstances, prior map $M_i$ has a large value for the pixels which include real changes and has a small value for the pixels in which only illumination variations happen. Since, illumination changes should be contained in a subspace orthogonal to the real changes, each input frame should satisfy the following constraints \cite{shakeri2017moving}:
\begin{equation} \label{constraints}
\begin{split}
& M_i^T \mid C_i \mid = 0 ,\\
& (1-M_i)^T \mid F_i \mid = 0\\
& s.t.\quad S_i = F_i + C_i,
\end{split}
\end{equation}
where $C_i$ is the illumination change image and $F_i$ is the image denoting moving foreground objects. $C_i$ is an optimizable parameter in our framework (Fig. \ref{graph}). Based on the illumination constraints, We can write the $L_{decomp}$ loss term that is responsible for decomposing $S_i$ into $C_i$ and $F_i$ as (\ref{Decomp_loss}): 
\begin{equation} \label{Decomp_loss}
L_{decomp} = \sum_i M_i^T \mid C_i \mid + \sum_i (1-M_i)^T \mid F_i \mid.
\end{equation}

\subsection{End-to-end optimization} 

To prevent overfitting to noise, we apply $l2$ regularization, or in other words weight decay, on the parameters of the generative networks. $L_{reg}$ of the network is defined as follows.
\begin{equation} \label{Reg_loss}
L_{reg} = \lambda (1/2\lVert W^1 \rVert_2^2 + 1/2\lVert W^2 \rVert_2^2)
\end{equation}
$W^1$ and $W^2$ denote parameters of Net1 and Net2, except biases. $\lambda$ is a hyper-parameter. The overall loss function of the whole framework is defined in (\ref{loss}):
\begin{equation} \label{loss}
L  = L_{reconst} + L_{decomp} + L_{reg}
\end{equation}

For optimization, we perform no preprocessing on the input data. We optimize the network for each image sequence independently. As mentioned earlier, NUMOD does not use ground-truths in the loss function $L$ and it is optimized in an unsupervised manner. We use Adam \cite{kingma2014adam}, a stochastic gradient-based optimizer, to optimize all the parameters $\{U_i^1,U_i^2,C_i \}_{i=1}^n, W^1, W^2$. Due to the end-to-end optimization, Net1 and Net2 can have a good effect on each other and lead each other for a more accurate decomposition. 

Finally, we apply a threshold on $F_i$ vector at each pixel location $(x,y)$ to obtain foreground binary mask $b_i$.
\begin{equation} \label{thresholding}
b_i(x,y) =\begin{cases} 1 & \mid F_i(x,y) \mid \geq 2t \\
                   0 &  \mid F_i(x,y) \mid < 2t
       \end{cases}
\end{equation}
where $t$ is the standard deviation of pixels in $\{F_i\}_{i=1}^n$.

\subsection{Online mode}
The advantage of NUMOD is that it works in both batch and online modes. In the batch mode, like other batch methods, we optimize the parameters on batches of the sequence of images. The extension to online mode is natural by the virtue of parameterized networks, Net1 and Net2.

For the online mode, first, we optimize the network on an initial batch of images. Then, parameters $W^1$ and $W^2$ of the networks are frozen and the other variables $U^1_i$, $U^2_i$ and $C_i$ are left optimizable (Fig. \ref{graph}). We freeze network parameters to avoid overfitting while fine-tuning NUMOD for the oncoming stream of image frames. At this point, one or a few input frames are feed into the network, incrementally, and by backpropagating the error, their corresponding low-dimensional latent variables $U^1_i$, $U^2_i$ and illumination variable $C_i$ are optimized. The loss term dose not change during online mode except that $L_{reg}$ is omitted:
\begin{equation} \label{online_loss}
L_{online}  = L_{reconst} + L_{decomp}. 
\end{equation}
Based on our experimental results, with an adequate size of the initial batch, the network parameters are well-optimized in the initial phase, so that after freezing those and feeding new frames, and just by optimizing $U^1_i$, $U^2_i$ and $C_i$, NUMOD is able to learn accurate background images and separate real changes from shadows and illumination changes.
\section{experimental results and discussion}

\subsection{Experimental Setup}
\subsubsection{Benchmark data sets}
We evaluate our proposed method on some benchmark data sets that include illumination changes and moving shadows to demonstrate that NUMOD can handle illumination changes.

Image sequences ``LightSwitch'' and ``Camouflage'' from Wallflower data set \cite{toyama1999wallflower} and ``Lobby'' from I2R dataset \cite{li2004statistical} are selected due to sudden and global illumination changes. We Also selected ``Cubicle'', ``PeopleInShade'', ``CopyMachine'' and ``Backdoor'' sequences from CDnet data set \cite{goyette2012changedetection} that include illumination changes and moving shadows and contain both indoor and outdoor scenes. 
\subsubsection{Evaluation metric}
We use F-Measure metric for comparing different algorithms, which is used generally as a overall performance indicator of the moving object detection and background subtraction methods. F-measure is defined as follows.
\begin{equation} \label{F-Measure}
\text{F-measure}  = 2* \frac {\text{Recall}*\text{Precision}}{\text{Recall}+\text{Precision}}
\end{equation}
For each sequence, F-measure is computed for each individual frame first and then the average over all the frames are computed and reported.
\subsubsection{Network setup}
The size of latent vectors $U^1_i$ and $U^2_i$ is $5$. The first and second layer size are $10$ and $20$, respectively. The only hyperparameter of the network is $\lambda$ in (\ref{Reg_loss}), which is set $0.005$ in all our experiments. Adam optimizer algorithm \cite{kingma2014adam} with fixed learning rate of $0.001$ is used throughout. For the online mode, the first half of the sequence is used to pre-train the network and then streams of $10$ frames are used for online optimization.

\subsection{Evaluation of NUMOD batch mode}
\begin{table*}
\begin{center}
  \label{tab:batch}
  \begin{tabular}{lccccccccc}
    \hline
    Sequence&SSGoDec&PRMF&Decolor&PCP&BRMF&LSD&ILISD&BEN-BLN&NUMOD-Batch\\
    &\cite{zhou2011godec}&\cite{wang2012probabilistic}&\cite{zhou2013moving}&\cite{candes2011robust}&\cite{wang2013bayesian}&\cite{liu2015background}&\cite{shakeri2017moving}&\cite{xu2014dynamic}&\\    
    \hline
    Backdoor&0.6611&0.7251&0.7656&0.7594&0.6291&0.7603&0.8150&0.6212&\textbf{0.8536}\\
    CopyMachine&0.5401&0.6834&0.7511&0.6798&0.3293&0.8174&0.8179&0.5518&\textbf{0.8519}\\
    Cubicle&0.3035&0.3397&0.5503&0.4978&0.3746&0.4232&0.6887&0.4868&\textbf{0.7468}\\
    PeopleInShade&0.2258&0.5163&0.5559&0.6583&0.3313&0.6168&0.8010&0.6802&\textbf{0.8661}\\
    Camouflage&0.6452&0.6048&0.8125&0.3388&0.6048&0.9456&0.8663&\textbf{0.9552}&0.9224\\
    LightSwitch&0.3804&0.2922&0.5782&\textbf{0.8375}&0.2872&0.6640&0.7128&0.0868&0.7761\\
    Lobby&0.0831&0.6256&0.7983&0.6240&0.3161&0.7313&0.7849&0.4691&\textbf{0.8355}\\
  \hline
\end{tabular}
\caption{Performance comparison of batch methods based on F-measure score}
\end{center}
\end{table*}

\begin{figure}
\begin{center}
\includegraphics[width=0.7\columnwidth]{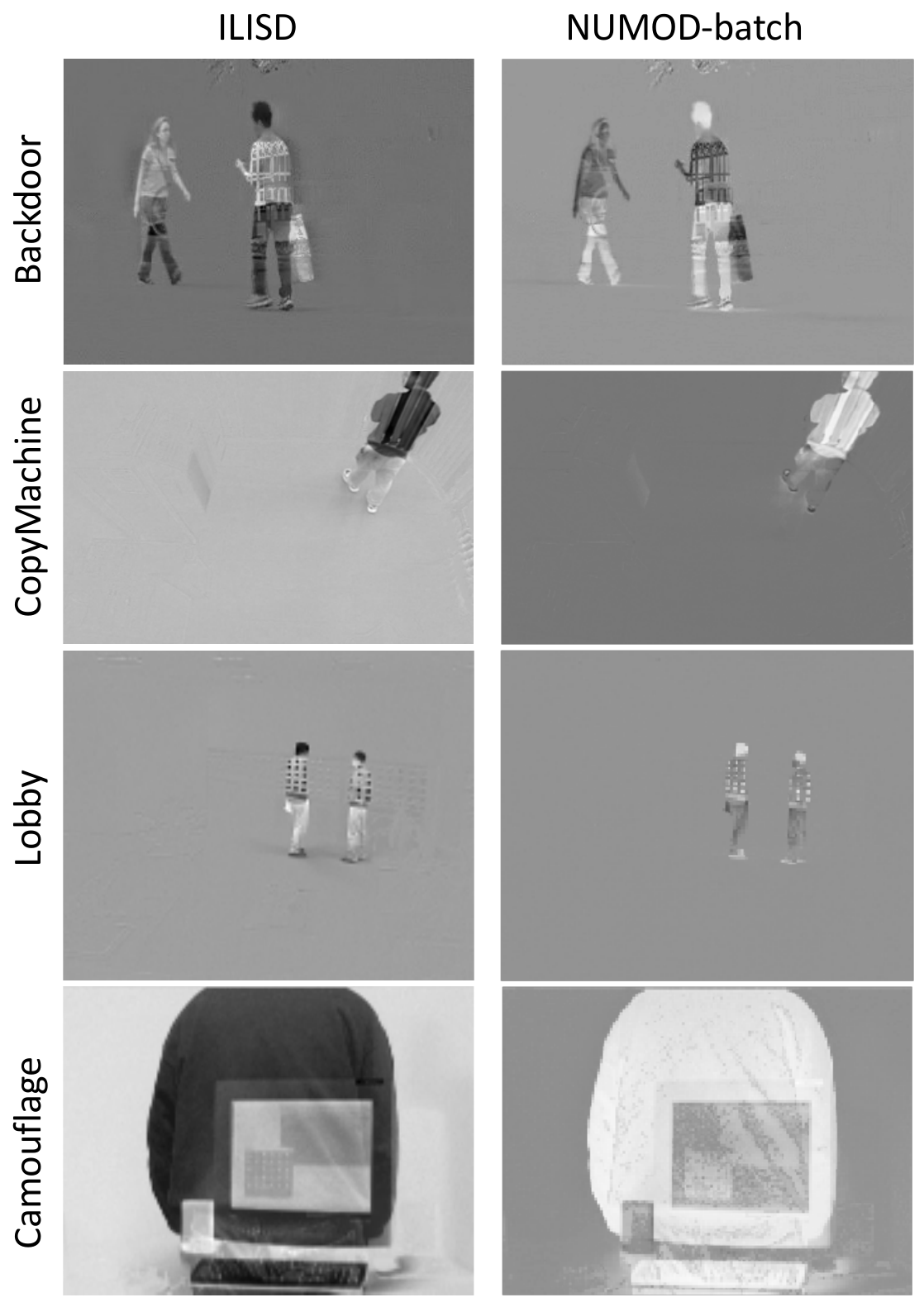}
\caption{Comparison of ILISD and NUMOD-batch}
\label{fig:ILISD_NUMOD}
\end{center}
\end{figure}

In the first set of experiments, we compare batch mode of NUMOD to competing batch methods. These methods include some of the best low-rank and sparse decomposition methods SSGoDec\cite{zhou2011godec}, PRMF\cite{wang2012probabilistic}, Decolor\cite{zhou2013moving}, PCP\cite{candes2011robust}, BRMF\cite{wang2013bayesian}, LSD\cite{liu2015background} and ILISD\cite{shakeri2017moving}. ILISD is the method that can handle illumination change and moving shadows and comes closest to NUMOD in terms of the objective function. We also compare our method to BEN-BLN \cite{xu2014dynamic}, which is based on stacked autoencoders.

Performance comparison of batch methods is presented in table~\ref{tab:batch} in terms of F-measure. We observe that NUMOD outperforms other methods in most of the sequences. In ``Camouflage'' sequence, a person suddenly appears in a large portion of the scene and causes illumination change. For ``LightSwitch'' and ``Camouflage'' sequences, ground-truth is available just for one frame and so the results are not based on the performance over the whole sequence. Generally, the results demonstrate capability of our method in separating illumination and real moving objects.

In Fig. \ref{fig:ILISD_NUMOD} we show samples for comparing qualitative results between ILISD and NUMOD-batch. The shown results are the decomposed foreground part, called $F_i$ in our notations, in both methods. Since ILISD result is in grayscale, we also show our result in grayscale to make comparison easier. We selected ILISD for comparison since it is the only batch method capable of handling severe illumination changes and comes closest to NUMOD. As it can be seen, these methods produce comparable results.

\subsection{Evaluation of NUMOD online mode} 
\begin{table*}
\begin{center}
  \label{tab:online}
  \begin{tabular}{lcccccc}
    \hline
    Sequence&GMM\cite{stauffer1999adaptive}&GRASTA\cite{he2012incremental}&OR-PCA\cite{feng2013online}&COROLA\cite{shakeri2016corola}&BEN-BLN-Online\cite{xu2014dynamic}&NUMOD-Online\\
    \hline
    Backdoor&0.6512&0.6822&0.7360&0.6821&0.5539&\textbf{0.8394}\\
    CopyMachine&0.5298&0.6490&0.5599&0.4155&0.5197&\textbf{0.8589}\\
    Cubicle&0.3410&0.4113&0.5998&0.5213&0.4164&\textbf{0.7473}\\
    PeopleInShade&0.3305&0.5288&0.6088&	
0.2474&0.7380&\textbf{0.8671}\\
    Camouflage&0.8102&0.6528&0.0823&0.7138&0.8311&\textbf{0.9117}\\
    LightSwitch&0.4946&0.5631&\textbf{0.8006}&0.2830&0.2707&0.7518\\
    Lobby&0.3441&0.6727&0.5831&0.7641&0.3380&\textbf{0.7687}\\
  \hline
\end{tabular}
\caption{Performance comparison of online methods based on F-measure score}
\end{center}
\end{table*}

\begin{figure}
\begin{center}
\includegraphics[width=\columnwidth]{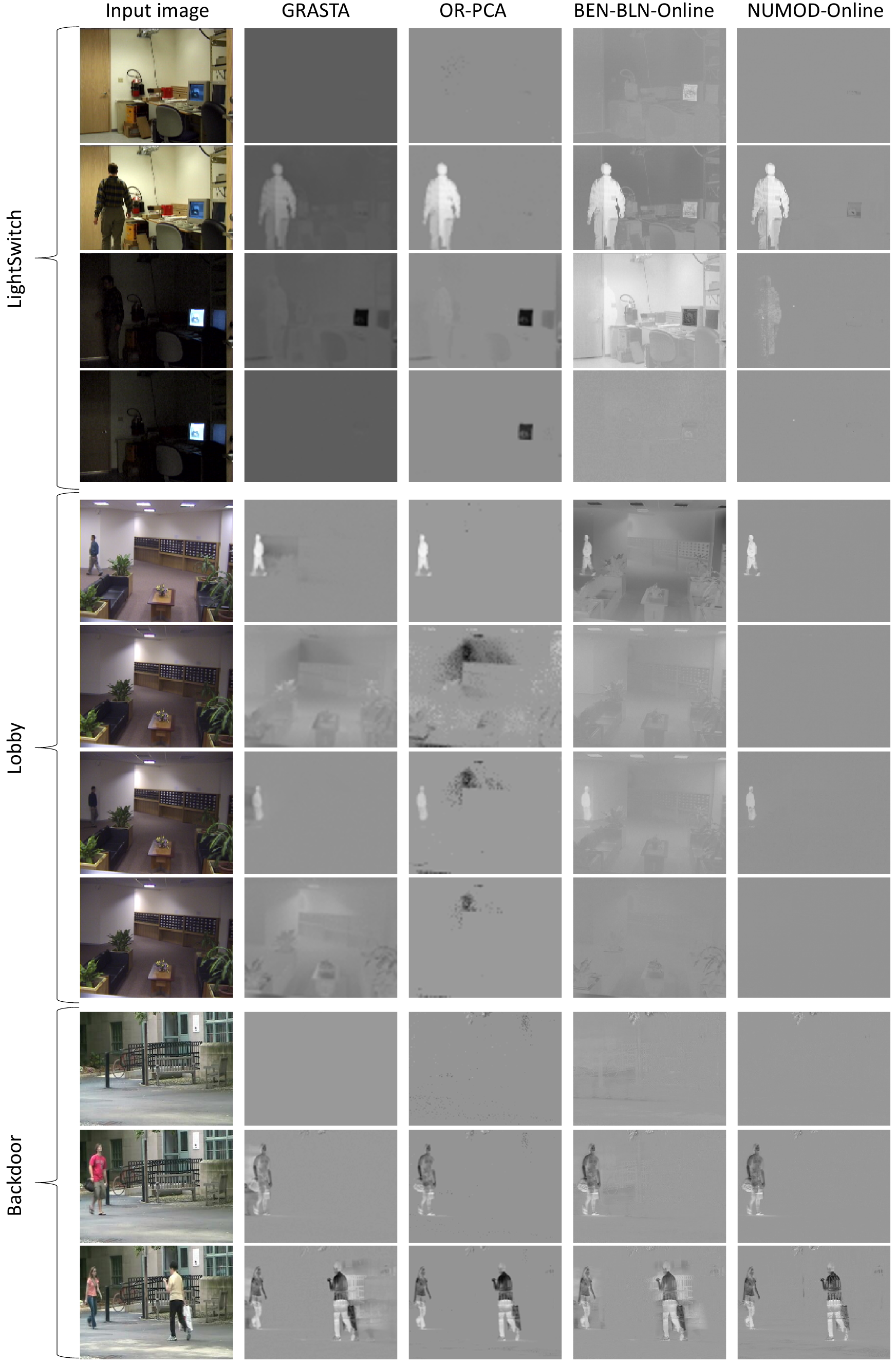}
\caption{Qualitative results for online methods on ``LightSwitch'', ``Lobby'' and ``Backdoor'' sequences. Columns from left to right are input image, sparse output of GRASTA, OR-PCS and BEN-BLN-Online algorithms and moving foreground output,$F_i$, of NuMOD-online in grayscale, respectively.}
\label{fig:qualitative_online}
\end{center}
\end{figure}

In this set of experiments, we compare online mode of NUMOD to competing online/sequential methods including the GMM method \cite{stauffer1999adaptive} and some of the best LRS decomposition methods GRASTA\cite{he2012incremental}, OR-PCA\cite{feng2013online}, COROLA\cite{shakeri2016corola} and online mode of BEN-BLN \cite{xu2014dynamic} neural network based algorithm \cite{xu2014dynamic}.

Performance of each method based on F-measure is reported in table \ref{tab:online}. Our proposed method obtains the best results in all sequences except ``LightSwitch''. As we mentioned earlier, results for this sequence are based on only a single ground-truth frame. Numerical results shows that our method outperforms other online methods and is able to handle illumination changes.

In Fig. \ref{fig:qualitative_online} we compare qualitative results of online methods on ``LightSwitch'', ``Lobby'' and ``Backdoor'' sequences. The results from other methods is their sparse output and the result of our method is the the moving foreground output, $F_i$ in grayscale. In the ``LightSwitch'' sequence there is a sudden and heavy illumination change. The third row shows all other methods fail to handle illumination change except NUMOD. The first and fourth rows demonstrates that NUMOD is able to obtain a illumination-free foreground when no moving object is in the scene. 

The ``Lobby'' sequence has a moderate illumination change. in this sequence, referring to Fig. \ref{fig:qualitative_online}, the only method that is able to distinguish moving foreground and illumination changes is NUMOD. GRASTA and OR-PCA are doing a good job in some frames but in other frames their foreground contains illumination changes. BEN-BLN is not able to capture the illumination changes in its background and as it is shown it appears in the foreground. Once more, we see in ``Lobby'' sequence results that only NUMOD's foreground does not include illumination changes in frames that do not have any moving objects. 

The last three rows in Fig. \ref{fig:qualitative_online} shows results for ``Backdoor'' sequence. In this sequence, OR-PCA and NUMOD are doing a good job. However, in the frame without moving objects, OR-PCA captures some noises in its result.

In general, the quantitative and qualitative results demonstrate that NOMOD online mode can outperforms other methods in handling illumination changes. 

\section{Conclusions and Future Work}
In this paper, we proposed a MOD method called NUMOD based on a generative neural network model. In NUMOD, unsupervised generative networks learn low-dimensional latent representations of the high dimensional background images. For separating illumination changes from foreground moving objects, NUMOD uses an illumination invariant representation of images. The method works well both in the online and the batch modes. To the best of our knowledge, NUMUD is the only online method that can handle illumination changes quite successfully. Based on the quantitative and qualitative experimental results, our method outperforms batch and online methods in most of the bench mark image sequences.

In the current work, we did not use any assumption considering structure of the sparse foreground regions. For the future work, we want to enhance the results by considering the connectivity of foreground outliers.

{\small
\bibliographystyle{ieee}
\bibliography{ref}
}

\end{document}